\documentclass[
]{ceurart}

\usepackage{listings}
\begin{document}

%%
%% Rights management information.
%% CC-BY is default license.
\copyrightyear{2025}
\copyrightclause{Copyright for this paper by its authors.
  Use permitted under Creative Commons License Attribution 4.0
  International (CC BY 4.0).}

%%
%% This command is for the conference information
\conference{Challenge and Workshop (BC9): Large Language Models for Clinical and Biomedical NLP, International Joint Conference on Artificial Intelligence (IJCAI), August 16–22, 2025, Montreal, Canada}

%%
%% The "title" command
\title{FMI@SU ToxHabits: Evaluating LLMs Performance on Toxic Habit Extraction in Spanish Clinical Texts}

%\tnotemark[1]
%\tnotetext[1]{You can use this document as the template for preparing your
%  publication. We recommend using the latest version of the ceurart style.}

%%
%% The "author" command and its associated commands are used to define
%% the authors and their affiliations.
\author[1]{Sylvia Vassileva}[%
email=svasileva@fmi.uni-sofia.bg,
]
\cormark[1]
%\address[1]{Faculty of Mathematics and Informatics, Sofia University St. Kliment Ohridski, Sofia, Bulgaria}

\author[1] {Ivan Koychev}[%
email=koychev@fmi.uni-sofia.bg,
]
%\address[1]{Faculty of Mathematics and Informatics, Sofia University St. Kliment Ohridski, Sofia, Bulgaria}

\author[1, 2]{Svetla Boytcheva}[%
email=svetla.boytcheva@graphwise.ai,
]
\address[1]{Faculty of Mathematics and Informatics, Sofia University St. Kliment Ohridski, Sofia, Bulgaria}
\address[2]{Graphwise, Sofia, Bulgaria}

%\author[1,2]{Dmitry S. Kulyabov}[%
%orcid=0000-0002-0877-7063,
%email=kulyabov-ds@rudn.ru,
%url=https://yamadharma.github.io/,
%]
%\cormark[1]
%\fnmark[1]

%% Footnotes
\cortext[1]{Corresponding author.}
%\fntext[1]{These authors contributed equally.}

%%
%% The abstract is a short summary of the work to be presented in the
%% article.
\begin{abstract}
The paper presents an approach for the recognition of toxic habits named entities in Spanish clinical texts. The approach was developed for the ToxHabits Shared Task. Our team participated in subtask 1, which aims to detect substance use and abuse mentions in clinical case reports and classify them in four categories (Tobacco, Alcohol, Cannabis, and Drug). We explored various methods of utilizing LLMs for the task, including zero-shot, few-shot, and prompt optimization, and found that GPT-4.1's few-shot prompting performed the best in our experiments. Our method achieved an F1 score of 0.65 on the test set, demonstrating a promising result for recognizing named entities in languages other than English.
\end{abstract}

%%
%% Keywords. The author(s) should pick words that accurately describe
%% the work being presented. Separate the keywords with commas.

%%
%% This command processes the author and affiliation and title
%% information and builds the first part of the formatted document.
\maketitle

\section{Introduction}
% Sylvia
There is a vast amount of clinical information in case reports and discharge summaries, which is more challenging to process due to their lack of structure. It is estimated that 80\% of clinical data exists in unstructured format, making it harder to use by automatic methods of analysis \cite{unstructuredbigdata}. Therefore, the ability to extract structured information from clinical texts can help advance clinical research and inform patient care of similar cases and outcomes.
About 296 million people aged 15-64 years used psychoactive drugs in 2021, which makes substance use and abuse a significant issue for population health worldwide. Substance abuse increases morbidity and mortality risks and adversely impacts society as a whole \footnote{\url{https://www.who.int/health-topics/drugs-psychoactive\#tab=tab\_2}}. Mining Electronic Health Records for different types of information related to drug use and abuse can help provide structured data for clinical research and automated analysis of drug use cases, informing future drug therapy regimens. Bearing this in mind, building systems that automatically detect and extract drug-related information from clinical texts is an important step in better understanding toxic habits.

ToxHabits \footnote{\url{https://temu.bsc.es/toxhabits/}} is a shared task and dataset focusing on Named Entity Recognition (NER) of toxic substance use and abuse in clinical texts in Spanish \cite{toxhabitsoverview}. ToxHabits is a part of the BioCreative IX Workshop.
The shared task consists of two subtasks:
\begin{itemize}
    \item  Subtask 1 - toxic habits NER - detecting substance use and abuse in Spanish clinical case reports; each mention referring to toxic habits is called a trigger phrase;
    \item Subtask 2 - detection of different characteristics of the toxic triggers, like substance type (cocaine, heroine, etc.), method (IV, inhaled), amount (2 drinks), frequency (every day), duration (for two years), history (stopped using in 2007), and status time - current, past, etc.
\end{itemize}

Due to time constraints, our team participated only in the first subtask. Our code and prompts will be publicly available in GitHub\footnote{\url{https://github.com/svassileva/fmi-toxhabits-ner}}.

\section{Related Work}
% Svetla

Following the recent advancements in generative AI models, many researchers have explored the capabilities of LLMs for NER in biomedical texts. This dramatically changed how the NER task is tackled, with the development of labeling approaches (primarily dictionary-based, rule-based, or supervised deep learning) to natural language generation task employing either unsupervised techniques like zero-shot prompting or providing few-shot fine tuning \cite{10.1093/jamia/ocae202}, \cite{hu2024improving}. Another innovative solution proposed is the Retrieval Augmented Generation (RAG) application \cite{monajatipoor2024llms}, which includes various hybrid methods for enhancement, such as external knowledge incorporation ~\cite{bian2023inspire} and dictionary-based approaches ~\cite{garcia2025nssc}. There are also specialized LLMs developed with domain adaptations, such as Llama2-MedTuned~\cite{ROHANIAN2024103007}, and instruction fine-tuned models for task adaptation, like UniversalNER \cite{zhou2023universalner}. Although significant efforts have been made in this direction, LLMs continue to struggle with solving the NER task for biomedical texts in general. There are two aspects to the NER task: identifying the start and end indices of the named entity and assigning a category. The first subtask is too challenging for LLMs; they can identify the entity, but are not so precise in the start-end indices ~\cite{lu2025large}. This is manageable with some post-processing techniques, such as the chain of thoughts (COT) ~\cite{bian2023inspire} or even simple scripts. For the second subtask, the LLMs present outstanding capabilities. Some authors even enhance the LLM performance by incorporating external knowledge, such as UMLS ~\cite{Bodenreider2004-ub} and category voting for predictions.  
Some promising results are achieved partially for some specific types of entities - F1 score 0.72 for "Species" and 0.69 for "gene/protein" ~\cite{cohen2017colorado}, F1-score of 0.94 for the categories "chemical" and "species" ~\cite{biana2024vaner}.

\section{Data}
The ToxHabits dataset consists of 1,500 clinical case reports in Spanish from various medical specialties sourced from medical journals \cite{toxhabitsner}.
The corpus is labeled with mentions of toxic habits - substance use and abuse, as well as attributes of the specific habit.
Each toxic habit mentioned, called a trigger, was classified according to its type - tobacco, cannabis, alcohol, or drug. Different attribute types are labeled for each trigger, including the substance type, method of usage, amount, frequency, duration, history, and status time.
The train set consisted of 1,200 documents, while the test set had 300 documents. We split the train set further and allocated 300 documents for our validation/dev set, which we use for evaluating the performance of different models before submission. The statistics of the dataset for subtask 1 are shown in Table \ref{tab:ner_dataset_stats}. The distribution of labels per type in the train and dev sets is shown in Table \ref{tab:ner_dataset_stats_per_type}. The Drug type is predominant, with a lot of different examples of drug mentions.

% Sylvia

\section{Methods}
We participated in the first subtask only - Named Entity Recognition of toxic habit trigger mentions.
We investigated various approaches involving OpenAI LLMs and found that the best-performing model was GPT-4.1, utilizing few-shot prompting.

The method we propose relies on the in-context learning capabilities of LLMs and consists of the following steps:
\begin{enumerate}
    \item The case report is split into sections (based on two subsequent new line characters); we found that the model retrieves a higher number of correct entities when the input text is shorter, when testing with a small subset of the data.
    \item A random set of 5 examples is sampled from the train set to use for in-context learning;
    \item The prompt is constructed using the few-shot examples and a prompt explaining the task; we utilize the model's structured output capabilities to extract phrases containing toxic habits for each type: tobacco, alcohol, cannabis, and drugs.
    \item The mention spans are retrieved by searching for the extracted phrases in the text; if there is an overlap of mentions, we select the shorter span, as we noticed that the labeled mentions in the train set tend to be shorter.
\end{enumerate}

\begin{figure*}[h!]
	\centering
	\includegraphics[width=0.85\textwidth]{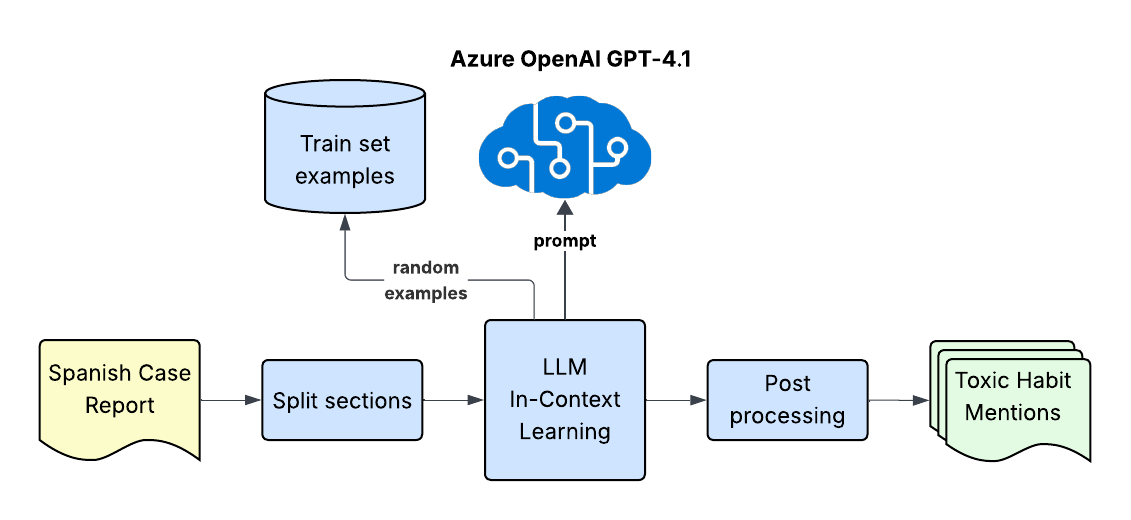}
	\caption{The process for mention extraction using an LLM.}
	\label{fig:method}
\end{figure*}

The process is represented in Fig. \ref{fig:method}. We use the DsPy library\footnote{\url{https://dspy.ai/}} \cite{khattab2024dspy} to perform our experiments. We used the default temperature - 0, top\_p - 1, max tokens - 4000.

\begin{table}[h!]
\centering
\begin{tabular}{@{}lrrr@{}}
\toprule
\textbf{Characteristic} & 
\textbf{Train} & 
\textbf{Dev} & 
\textbf{Test} \\
\midrule
\# Mentions & 5,754 & 1,838 & - \\
% unique mentions
\# Documents & 900 & 300 & 300 \\
\# Sentences & 28,971 & 8,939 & 8,626 \\
\bottomrule
\end{tabular}
\caption{Dataset statistics for substask 1.}
\label{tab:ner_dataset_stats}
\end{table}

\begin{table}[h!]
\centering
\begin{tabular}{@{}lrrrr@{}}
\toprule
\textbf{Dataset} & 
\textbf{Tobacco} & 
\textbf{Alcohol} & 
\textbf{Cannabis} &
\textbf{Drug} \\
\midrule
Train set & 640 & 948 & 599 & 3,567 \\
Dev set & 193 & 332 & 194 & 1,119 \\
% unique mentions per type
\bottomrule
\end{tabular}
\caption{Trigger mentions per type on train and dev sets.}
\label{tab:ner_dataset_stats_per_type}
\end{table}
% Sylvia

\section{Experiments and Results}
% Sylvia
As a baseline approach, we created a dictionary from the train set and filtered the mentions that were unambiguously labeled as entities. Also, we trained a BERT-based model on token classification (Spanish Clinical RoBERTa), which has shown good results on Spanish clinical named entity recognition.

We performed some preliminary small experiments with several OpenAI models, including GPT-4o, GPT-4.1, and GPT-4.5 preview, and GPT-4.1 showed the best initial results. Therefore, for our experiments, we focused on GPT-4.1 using a private Azure OpenAI deployment.
We compared the performance of GPT-4.1 in zero-shot and few-shot settings on the dev set and found that few-shot outperforms zero-shot by about 6\% F1 and the dictionary baseline by about 2\%. In the few-shot setting, we used 3, 5, or 7 examples randomly sampled from the training set. We used the default temperature - 0, top\_p - 1, and max tokens - 4000. Using three examples underperformed; however, there was a very small difference between using 5 and 7 examples, and we chose to proceed with 5 to reduce the number of tokens used and thus the cost per document.

We attempted to run the kNN Few-shot DsPy optimizer, which selects examples based on semantic similarity with each record during inference. However, due to time constraints, we were unable to complete the experiment. As future work, it would be beneficial to explore this approach, as kNN Few-Shot has shown improvement in results in cases where high-quality examples are crucial \cite{sarmah2024comparativestudydspyteleprompter}. 

After the submission phase, we performed additional experiments with DsPy Multiprompt Instruction Proposal Optimizer v2 (MIPROv2) \footnote{\url{https://dspy.ai/api/optimizers/MIPROv2/}}, which improved the performance on the dev set by about 1\%. For this experiment, we also used the default temperature - 0, top\_p - 1, and max tokens - 4000. We optimized the Intersection over Union (IoU) metric at the character level to provide the model with more granular feedback on its proximity to the correct mention.

Also, we tried to optimize the prompt manually and were able to improve the precision and F1 score by approximately 1\% higher than the automatically optimized one. During the experiment with the manually optimized prompt, we had to increase the max token length several times, and it reached 16,000 max tokens. The temperature for that experiment was set to 1, so that would explain the response verbosity. The value for top\_p was also 1, and the max retries were set to 2. 
Details about the prompt and structured extraction model are available on our GitHub.
Evaluation results from our experiments on the dev set are shown in Table \ref{tab:ner_dev_results}.

We submitted several different models and hybrid combinations to the competition, including a dictionary baseline, GPT-4.1 zero and few-shot, as well as the combination of dictionary and GPT-based approaches. The results on the test set are shown in Table \ref{tab:ner_test_results}. Our best model was the GPT-4.1 few-shot model, which achieved a precision of 0.59 and an F1 score of 0.65. However, the best recall is achieved by the Dictionary model. Unfortunately, the combination of the dictionary and the GPT model decreases the performance, as the combination strategy proved unsuccessful.

We also evaluated the models using customized metrics to determine what part of the gold entities is retrieved by the LLM as a partial match. We measure two different metrics - the ratio of gold entities contained completely in a predicted entity over the total number of gold entities (GC) and the ratio of gold entities contained completely in a predicted entity and matching the correct type over the total number of gold entities (GCT).
%; the ratio of gold entities partially overlapping with a predicted entity over the total number of gold entities (GO); 
%for each of the former metrics we also calculated the corresponding value but including a type check to ensure the predicted entity has the same type as the gold one with which it overlaps. 
The results of the evaluation a presented in Table \ref{tab:ner_dev_results_overlaps}. The highest-scoring models are the combined dictionary and GPT-4.1 models (zero and few-shot). The standalone GPT-4.1 zero-shot shows competitive results on these metrics, similar to the Spanish Clinical RoBERTa. GPT-4.5 preview seems to make a lot more mistakes in the entity type, as the type-specific metrics show lower scores than all other models. If a method is developed to refine the predicted entities from the LLM model, it can potentially increase the recall performance of the system.

\begin{table}[h!]
\centering
\begin{tabular}{@{}lccc@{}}
\toprule
\textbf{Model} & 
\textbf{Precision} & 
\textbf{Recall} & 
\textbf{F1} \\
\midrule
Dictionary Baseline & 0.4936 &	\textbf{0.8270} &	0.6182 \\
Spanish Clinical RoBERTa  &  \textbf{0.7885} & 0.7933 & \textbf{0.7909} \\
GPT-4.1 zero-shot &	 0.4942	& 0.6860	& 0.5745 \\
GPT-4.1 few-shot (3 examples) &	0.5064 &	0.6766 &	0.5793 \\
GPT-4.1 few-shot (5 examples) &	 0.5687	& 0.7181 &	0.6347 \\
GPT-4.1 few-shot (7 examples) &	 0.5712 &	0.7247 &	0.6389 \\
GPT-4.1 few-shot DsPy optimizer &	 0.5706	& 0.7457 &	0.6465 \\
GTP-4.1 few-shot optimized prompt & \underline{0.5938} &	0.7242 &	\underline{0.6526} \\
GPT-4.5 preview & 0.3765 &	0.5196 &	0.4366 \\
Dictionary+GPT-4.1 zero-shot &	0.4666 &	0.8264	& 0.5964  \\
Dictionary+GPT-4.1 few-shot &	0.4653 &	\underline{0.8336}	& 0.5972  \\
\bottomrule
\end{tabular}
\caption{Results of the models for the NER task on the dev set.}
\label{tab:ner_dev_results}
\end{table}

\begin{table}[h!]
\centering
\begin{tabular}{@{}lcc@{}}
\toprule
\textbf{Model} & 
\textbf{GC} & 
\textbf{GCT} \\
%\textbf{GO} &
%\textbf{GOT}\\
\midrule
Dictionary Baseline & 0.8374 & 0.8302 \\
%& 0.85903 & 0.8485 \\
Spanish Clinical RoBERTa & 0.8203 & 0.8109 \\
%& 0.8540 & 0.8435\\
GPT-4.1 zero-shot & 0.8308 & 0.8126 \\
%& 0.8468 & 0.8286\\
GPT-4.1 few-shot (3 examples) & 0.8076 & 0.7888 \\
%& 0.8291 & 0.8103 \\
GPT-4.1 few-shot (5 examples) & 0.7888 & 0.7716 \\
%& 0.8114 & 0.7938\\
GPT-4.1 few-shot (7 examples) & 0.7960 & 0.7783 \\
%& 0.8181 & 0.7998 \\
GPT-4.1 few-shot DsPy optimizer & 0.8098 & 0.8009 \\
%& 0.8308 & 0.8214\\
%GTP-4.1 few-shot optimized prompt & & & & \\
GPT-4.5 preview & 0.8385 & 0.5550 \\
%& 0.8601 & 0.5743 \\
Dictionary+GPT-4.1 zero-shot & \textbf{0.8584} & \textbf{0.8413} \\
%& \textbf{0.9496} & \textbf{0.9281} \\
Dictionary+GPT-4.1 few-shot & \underline{0.8479} & \underline{0.8319} \\
%& \underline{0.9441} & \underline{0.9242} \\
\bottomrule
\end{tabular}
\caption{Evaluation of the models on the dev set on metrics reflecting the overlap with the gold entities: \textbf{GC} - \textbf{G}old entity is \textbf{C}ontained in a predicted entity, \textbf{GCT} - \textbf{G}old entity is \textbf{C}ontained in a predicted entity and has the correct \textbf{T}ype.}
\label{tab:ner_dev_results_overlaps}
\end{table}
%, \textbf{GO} - \textbf{G}old entity has partial \textbf{O}verlap with a predicted entity, \textbf{GOT} - \textbf{G}old entity has partial \textbf{O}verlap with a predicted entity and is of the correct \textbf{T}ype

\begin{table}[h!]
\centering
\begin{tabular}{@{}lccc@{}}
\toprule
\textbf{Model} & 
\textbf{Precision} & 
\textbf{Recall} & 
\textbf{F1} \\
\midrule
Dictionary Baseline &	0.46	& \textbf{0.87} &	0.6 \\
GPT-4.1 zero-shot &	0.49	& 0.68 &	0.57 \\
GPT-4.1 few-shot &	\textbf{0.59}	& 0.72	& \textbf{0.65} \\
Dictionary+GPT-4.1 zero-shot &	0.43 &	0.83	& 0.57 \\
Dictionary+GPT-4.1 few-shot &	0.44 &	0.83	& 0.57 \\
\bottomrule
\end{tabular}
\caption{Results of the submitted models for the NER task on the test set.}
\label{tab:ner_test_results}
\end{table}
% example

\section{Discussion}
% Sylvia
% error analysis
The most significant source of errors in the GPT-4.1 predictions is the correct detection of entity boundaries. Very often, the correct mention is part of a phrase extracted by the model, but the model retrieves additional context information related to the trigger attributes, such as method, current/past usage, amount, etc. In the case of GPT-4.1 five-shot, the gold entity is contained in a predicted entity in about 78\% of the gold entities. Other LLM models show even higher scores on this metric, up to 83\% for GPT-4.1 zero-shot.
The LLM adds negation particles to the phrase, which is a very important part in interpreting the entity. Table \ref{tab:ner_examples} shows a few examples of errors where the predicted mention contains the correct one. In the future, a method to fine-tune the entity boundaries can be applied as a post-processing step to improve the exact match of entities predicted by the approach.

\begin{table}[h!]
\centering
\begin{tabular}{@{}cc@{}}
\toprule
\textbf{Predicted phrase} & 
\textbf{Correct Phrase} \\
\midrule
\textbf{tabaquismo} activo &	 tabaquismo \\
\textbf{fumaba} mucho & fumaba \\
\textbf{cocaína} esnifada & cocaína \\
consumo abusivo de \textbf{alcohol} & alcohol \\
fuma \textbf{hachís} & hachís \\
ex \textbf{fumador} & fumador \\
No \textbf{fumador} & fumador \\
\bottomrule
\end{tabular}
\caption{Examples of errors in the predicted vs the correct phrase.}
\label{tab:ner_examples}
\end{table}

In our experiments, we attempted to optimize the GPT-4.1 prompt manually. We found that retrieving the assertion type of the toxic habit — i.e., assertion or negation — helped the model retrieve entities with higher precision and recall. Correctly processing negation is crucial in the overall task of clinical information extraction, and the model effectively combined both the extraction of the trigger and assertion classification. In the future, it is worth exploring combining the extraction of triggers and attributes together, as LLMs may be able to streamline the task instead of using separate models for each one.

The method we propose automatically detects multiple instances of the same phrase in the clinical report and marks them with the predicted trigger type. However, not all instances were labeled by human experts, which introduces another source of error in our model's predictions. Annotation is a complex task, and therefore, it is challenging to achieve high inter-annotator agreement consistently. Thus, some variance in the annotators' interpretation of the guidelines may have affected the labels. However, a more thorough evaluation can be done after the guidelines are made available.

Our proposed method, using the GPT-4.1 model, can retrieve more mentions of different types of alcohols. For example, mentions of cervezas (beers), ron (rum), or chupitos (shots) are identified by GPT-4.1 but are not labeled by human experts. In some cases, the gold labels include punctuation as part of the span ("alcohol:") while our proposed method would retrieve only the main word ("alcohol"). 
Another type of entity that GPT-4.1 retrieves more often are prescription drugs like clonazepam, diazepam, and oxicodona de formulación retardada (delayed-release oxycodone), which may not necessarily be associated with substance abuse as they can be prescribed for certain conditions. In general, our method retrieves more specific mentions of substances and struggles to identify common words that reference a toxic habit like consumidores (consumers), positivos (positives), and tóxicos (toxic). The toxic habit type referenced depends on the context and is more challenging for our proposed method to identify correctly. Filtering out the extra entities retrieved by the LLM could improve the overall performance of the system.

\section{Limitations}
Our experiments were performed with GPT-4.1 version 2025-04-14, and using other versions of the model may not produce similar results. We conducted our experiments exclusively with the ToxHabits dataset as part of the challenge; however, applying the same method to other datasets may not yield comparable performance due to dataset-specific differences. The performance of our proposed method depends on the prompt and examples selected; therefore, it is possible that a more effective prompt or example selection process could yield a higher F1 score using GPT-4.1.

% Sylvia

\section{Conclusion}
In this paper, we presented a method for clinical NER of toxic habits in Spanish clinical case reports. 
Our method, utilizing GPT-4.1 and in-context learning, achieved an F1 score of 0.65 on the test set.
Although the LLM-based named entity recognition does not perform as well as the fine-tuned BERT-based models, the results indicate a promising direction for further research, particularly in the area of multilingual clinical NER. An advantage of using LLMs is that they support multiple languages and can be instructed with very few examples to identify entities correctly. 
As future work, it would be beneficial to explore selecting semantically similar examples for each record (kNN few-shot) as it could improve the few-shot performance of LLMs. Furthermore, refining the entity boundaries based on the entities retrieved by the LLM could tackle the problem of longer sequences being retrieved by the LLM. Using a dedicated post-processing step that classifies the likelihood that the span is an entity could improve the precision of the overall system.

% Sylvia

%\section{Appendices}

%%
%% The acknowledgments section is defined using the "acknowledgments" environment
%% (and NOT an unnumbered section). This ensures the proper
%% identification of the section in the article metadata, and the
%% consistent spelling of the heading.
\begin{acknowledgments}
This work was partially supported by the European Union-NextGenerationEU, through the National Recovery and Resilience Plan of the Republic of Bulgaria [Grant Project No. BG-RRP-2.004-0008].
Part of this works is also supported by European Union’s Horizon research and innovation programme project RES-Q PLUS [Grant Agreement No. 101057603]. Views and opinions expressed are however those of the author only and do not necessarily reflect those of the European Union. Neither the European Union nor the granting authority can be held responsible for them.
\end{acknowledgments}

%% The declaration on generative AI comes in effect
%% in Janary 2025. See also
%% https://ceur-ws.org/GenAI/Policy.html
\section*{Declaration on Generative AI}
 During the preparation of this work, the author(s) used Grammarly in order to: Grammar and spelling check. After using these tool(s)/service(s), the author(s) reviewed and edited the content as needed and take(s) full responsibility for the publication’s content. 

%%
%% Define the bibliography file to be used
\bibliography{sample-ceur}

%%
%% If your work has an appendix, this is the place to put it.
\appendix

\end{document}